\documentclass{article}

\usepackage{arxiv}

\usepackage[utf8]{inputenc} 
\usepackage[T1]{fontenc}    

\usepackage[colorlinks]{hyperref}
\hypersetup{
  linkcolor=blue,
  citecolor=blue,
  anchorcolor=blue
}

\usepackage{url}            
\usepackage{booktabs}       
\usepackage{amsfonts}       
\usepackage{amssymb}
\usepackage{nicefrac}       
\usepackage{microtype}      
\usepackage{lipsum}

\usepackage{graphicx}

\usepackage{etoolbox}

\usepackage{tabularx}
\usepackage{multirow}
\usepackage{rotating}
\usepackage{array}

\newcolumntype{P}{>{\raggedright\arraybackslash}m{1.8cm}}%
\newcolumntype{L}{>{\raggedright\arraybackslash}p{3.2cm}}%

\title{Machine Learning Approaches For Motor Learning: A Short Review}

\author{
Baptiste Caramiaux\,$^{1}$, Jules Fran\c{c}oise\,$^{2}$, Wanyu Liu\,$^{3}$, T\'eo Sanchez\,$^{1}$, and Fr\'ed\'eric Bevilacqua\,$^{3}$\\
$^{1}$ Universit\'e Paris-Saclay, CNRS, Inria, LRI, France \\
$^{2}$ Universit\'e Paris-Saclay, CNRS, LIMSI, 91400, Orsay, France \\
$^{3}$ STMS IRCAM - CNRS - Sorbonne Université, France
}

\begin{document}
\maketitle

\begin{abstract}

Machine learning approaches have seen considerable applications in human movement \textit{modeling}, but remain limited for motor \textit{learning}. Motor learning requires accounting for motor variability, and poses new challenges as the algorithms need to be able to differentiate between new movements and variation of known ones. In this short review, we outline existing machine learning models for motor learning and their adaptation capabilities. We identify and describe three types of adaptation: \textit{Parameter adaptation} in probabilistic models, \textit{Transfer and meta-learning} in deep neural networks, and \textit{Planning adaptation} by reinforcement learning. To conclude, we discuss challenges for applying these models in the domain of motor learning support systems.

\end{abstract}


\section{Introduction}

The use of augmented feedback on movements enables the development of interactive systems designed to facilitate motor learning. Such systems, that we refer as \textit{motor learning support systems}, require capturing and processing movement data and returning augmented feedback to the users. These systems have primarily been investigated in rehabilitation (e.g. motor recovery after injury~\cite{kitago2013motor}), or in other forms of motor learning induced contexts, such as dance pedagogy~\cite{riviere2019capturing} or entertainment~\cite{anderson2013youmove}.

Motor learning support systems model human movements, taking into account the underlying learning mechanisms. While computational models have been proposed for simple forms of motor learning~\cite{emken2007motor}, modeling the processes at play in more complex skill learning remains challenging.
Motor learning usually refers to two types of mechanisms: motor adaptation and motor skill acquisition.
The former, motor adaptation, is the process by which the motor system adapts to perturbations in the environment \cite{wolpert2011principles}.
Adaptation tasks take place over a rather short time span (hours or days) and does not involve learning a new motor policy. The latter, motor skill acquisition, involves learning a new control policy, including novel movement patterns and shifts in speed-accuracy trade-offs \cite{diedrichsen2015motor, shmuelof2012motor}. Complex skills are typically learned over months or years \cite{yarrow2009inside, anders2008deliberate}.

The need for computational advances in motor learning research has recently been pointed out in the field of neurorehabilitation \cite{reinkensmeyer2016computational}. We believe that data-driven strategies, using machine learning, represent a complementary approach to analytical models of movement learning. Recent results in machine learning have shown impressive advances in movement modeling, such as action recognition or movement prediction \cite{rudenko2019human, santos2019artificial}. However, it is still difficult to apply such approaches to motor learning support systems. In particular, computational models must meet specific adaptation requirements in order to address the different variability mechanisms induced by motor adaptation and motor skill learning. These models have to account for both fine-grained changes in movement execution arising from motor adaptation mechanisms, and more radical changes in movement execution due to skill acquisition mechanisms.

We propose a short review of the adaptation capabilities of machine learning applied to movement modeling. The objective of this review is not to be exhaustive, but rather to provide an overview on recent publications on machine learning that we found significant for motor learning research. We believe that such an overview is currently missing and can offer novel research perspectives, targeting primarily researchers in the field of motor learning and behavioural sciences. In order to build the review presented in this paper, we focused on recent articles (typically less than 10 years). At the time of writing (end of 2019), we queried four online databases (Google Scholar, PubMed, Arxiv, ACM Digital Library) combining the following keywords: ``Human Movement'', ``Motor Model'', ``Modeling/Modelling'', ``Tracking'', ``Control'', ``Synthesis'', ``Movement Generation'', ``Movement Prediction'', ``On-line Learning'', ``Adaptation'', ``Gesture Recognition'', ``Deep Learning'', ``Imitation Learning''. We then compiled the papers in a spreadsheet and conducted a selection based on the type of model adaptations, the modeling technique, the field and the input data considered. We summarize the review in Table 1 and identify three adaptation categories in machine learning based human modeling: (1) \textbf{Parameter adaptation} in probabilistic models, (2) \textbf{Transfer and meta-learning} in deep neural networks, and (3) \textbf{Planning adaptation} by reinforcement learning. We present the selected papers according to the type of adaptation and discuss their use in motor learning research.

\begin{table}
\begin{center}
\tiny
\begin{tabularx}{\textwidth}{|P|p{2cm}|p{3.8cm}|L|X|}
\hline
\textbf{Type of adaptation} & \textbf{Models}  & \textbf{Application domains } & \textbf{Input data}  &  \textbf{Papers}  \\ \hline\hline

    & GMM & Gesture-based Interaction    & Gesture and      & \cite{franccoise2016soundguides} \\
    &     &                              & movement data    & \cite{sarasua2016machine} \\
    &     & Human-robot interaction      & Robot arm        & \cite{calinon2007learning} \\
Parameter    &     &                              &                  & \cite{calinon2016tutorial} \\ \cline{2-5}
adaptation    & One-shot HMM & Gesture-based Interaction & Gesture and movement data & \cite{franccoise2018motion} \\ \cline{2-5}
    & Incremental HMM & Human-robot interaction & 3D Motion Capture & \cite{kulic2008incremental, kulic2012incremental} \\ \cline{2-5}
    & Stylistic HMM & Movement synthesis & 3D Motion Capture & \cite{tilmanne2012stylistic} \\ \cline{2-5}
    & Particle filtering & Gesture-based Interaction & Gesture and movement data & \cite{caramiaux2015adaptive} \\ \hline\hline


 & Temporal  & Health, rehabilitation           &  Inertial sensors & \cite{Rad} \\
 & CNN       & Interactive movement   &  Motion capture data & \cite{Holden2016} \\
Transfer &           &  generation   &   &  \\
learning &           &  Movement analysis &  Videos and force measurements &  \cite{zecha2018convolutional}\\ \cline{2-5}
 & 2D CNN    & Gesture-based interaction    &  Photo reflective data  &  \cite{kikui2018intra}  \\
 &           &  Health \& rehabilitation    &  EMG data               &  \cite{cote2019deep}  \\ \cline{2-5}
 &  RNN      &  Human-robot interaction       &                 &  \cite{cheng2019human} \\
 &           & Human motion prediction & Motion capture data & \cite{martinez2017human} \\ \cline{2-5}
 & Recurrent Encoder-Decoder & Human motion prediction & Motion capture data & \cite{wang2019vred} \\ \hline


       &  CNN-LSTM &  Human-robot interaction & Robot arm trajectory &  \cite{duan2017one} \\
       &   &    & Raw pixels &  \cite{finn2017one} \\
Meta-  &   &    &            &  \cite{yu2018one} \\ \cline{2-5}
learning &   Recurrent Encoder-Decoder &  Movement generation & Motion capture data&  \cite{gui2018few} \\ \hline\hline


    &  IRL &  Human-robot interaction & Joint dynamics  &  \cite{kolter2008hierarchical} \\
    &    &    &    &  \cite{finn2016guided} \\ \cline{2-5}
Planning    &  GAIL &  Human-robot interaction & Joint Dynamics  &  \cite{ho2016generative} \\
adaptation    &    &   &  &  \cite{guo2018generative} \\
    &    &    &  &  \cite{hausman2017multi} \\
    &  &   & Raw pixels &  \cite{zhu2018reinforcement} \\ \cline{2-5}
&  VAE $+$ GAIL &  Human-robot interaction & Joint Dynamics &  \cite{wang2017robust} \\
 \hline

\end{tabularx}
\normalfont
\end{center}
\caption{Summary of the selected papers from our short survey, classified according to the type of adaptation involved in machine learning-based movement modeling.}
\label{summary_models}
\end{table}


\section{Parameter adaptation in probabilistic models}
\label{param_adaptation}

Research in movement recognition and generation has, for a long time, used parametric probabilistic approaches such as Gaussian Mixture Models (GMM), Hidden Markov Models (HMM), or Dynamic Bayesian Networks (DBN). These models are characterized by a set of trained parameters that can be adapted during execution, either by providing new examples along the interaction or adapting the model parameters online according to the characteristics of the task.

GMMs have been used in robotics and HCI to learn movement trajectory models from few demonstrations given by a human teacher \cite{calinon2007learning}. In robotics, \cite{calinon2016tutorial} proposed such an approach to adapt the robot movement parameters when new target coordinates are set for the robot arm. The underlying model is a GMM trained from few human movement demonstrations. In the context of movement-based interaction, \cite{franccoise2016soundguides} proposed a one-shot user adaptation process where the input movement associated with a sequence of sound synthesis parameters can be estimated from a single demonstration, in order to retrain the underlying GMM. They showed that user-adapted feedback can support the consistency of movement execution, but that the adaptation process is efficient for limited movement variations. \cite{sarasua2016machine} used GMM for soft recognition of conducting gestures that can adapt easily to user idiosyncrasies. The GMM-based mapping is learned from gesture demonstrations performed while listening to the desired musical rendering. The model is able to interpolate between demonstrations but cannot account for dramatic input variations. When tasks require to encode the dynamics and temporal evolution of the movement, generative sequence models such as HMMs have been applied to gesture recognition from few examples \cite{franccoise2018motion} as well as movement generation \cite{tilmanne2012stylistic}. Such adaptation techniques are often efficient when variations remain small in comparison with the overall movement dynamics.

Another approach, proposed by \cite{caramiaux2015adaptive} consists of tracking probability distribution parameters representing input movement variations from a set of gesture templates. Tracking uses particle filtering, which updates state parameters representing movement variations (such as scale, speed or orientation). The method can account for large slow variations. However, the tracking method does not learn the structure of the gesture variations and forgets previously observed states.

Finally, parametric probabilistic models can be trained online to account for new movement classes. \cite{kulic2008incremental,kulic2012incremental} proposed a HMM-based iterative training procedure for gesture recognition and generation.The method relies on unsupervised movement segmentation from which it automatically extracts existing and new primitives (using Kullback-Leibler divergence).This strategy enables both the fine-grained adaptation of existing motor primitives and the extension of the vocabulary of motor skills. However, unsupervised segmentation remains difficult for complex gestures, and the learning remains cumulative, with an ever growing vocabulary rather than a continuous adaptation to motor learning. Other online-strategies for segmentation with adaptive behaviour are described in \cite{KulicIEEE2009}.

In summary, parametric adaptation enables fine-grained adaptation to task variations and restricted input movement variations.
The typical use case is learning by demonstration (in human-robot interaction), or personalization (in human-machine interaction).

\section{Transfer and meta- learning in Deep Neural Networks}
\label{transfer_meta_learning}

Transfer and meta-learning are techniques aiming to accelerate and improve learning procedures of complex computational models such as Deep Neural Networks (DNN). The objective is to adapt pre-trained DNN efficiently to new tasks or application domains, unseen during training. This research is based on the literature in deep learning applied to movement modeling, which typically involves large datasets and benchmark-driven tests. The most popular approaches of this kind are Recurrent Neural Networks (RNN) \cite{fragkiadaki2015recurrent, mattos2015recurrent, alahi2016social, ghosh2017learning, martinez2017human, kratzer2019motion, wang2019vred}, and Temporal or Spatio-temporal Convolutional Neural Networks (CNN) \cite{li2018convolutional, gehring2017convolutional, li2019efficient, zecha2018convolutional}.

\subsection{Transfer learning}

Transfer learning adapts a pre-trained model on a source domain to new target tasks. Several strategies exist~\cite{Scott2018}. Transfer learning for movement modeling mainly relies on \textit{embedding learning}: movement features (or embeddings) are learned from the source domain, providing well-shaped features for the target domain.

Movement embeddings are learned from large movement datasets. A first strategy involves one-dimensional convolutions over the time domain \cite{Rad, Holden2016}. \cite{Rad} propose embedding learning using temporal convolution in order to improve diagnosis classification of autism spectrum disorder from intertial sensor data. The benefit of transfer learning is assessed on two datasets collected from the same participants, three years apart. In another context, \cite{Holden2016} makes use of transfer learning to synthesize movements from high-level control parameters easily configurable by human-users. Based on pre-trained movement embeddings from motion capture data, a mapping between high-level parameters and these embeddings can be efficiently learned according to the user needs.

Spatio-temporal convolutions can also be used to extract movement embeddings. \cite{kikui2018intra} uses this approach for inter- and intra-user adaptation of a gesture recognition system using photo reflective sensor data from a headset. They showed that transfer learning improves accuracy when the number of examples per class is low (lower than 6 ex/class). Also for classification, \cite{cote2019deep} showed that embedding learning systematically improved the classification accuracy of EMG-based movement data, in particular they found that embedding learning using CNNs on Continuous  Wavelet Transform (CWT) gives the best results.

Finally, RNN can also be used to learn movement embeddings, although this is not the most common approach. In the context of human-robot interaction, \cite{cheng2019human} proposed to train offline a RNN-based movement model and adapts the last layer parameters through recursive least square errors. The goal is to adapt the robot control command to human behaviour in real-time.

In summary, transfer learning of movement features has been proposed 1) to enable interactive movement generation or 2) to improve classification performance. Several problems remain to be addressed, especially in the context of motor learning. First, it is unclear how the model architecture and the size of the training set of the transfer task affects the approach. Second, it remains unexplored the extent to which successive transfers would provoke dramatic forgetting of previously transferred tasks.

\subsection{Meta-learning}

Meta-learning designates the ability of a system to \textit{learn how to learn} by being trained on a set of tasks (rather than a single task) such as learning faster (with fewer examples) on unseen tasks. Meta-learning is close to transfer learning, but, while transfer learning aims to use knowledge from a source application domain in order to improve or accelerate learning in a target application domain, meta-learning improves the learning procedure itself in order to handle various application domains.

Meta-learning of movement skills was proposed in robotics and human-robot interaction, to efficiently train robot actions from one or few demonstrations. \cite{duan2017one} proposed a one-shot imitation learning algorithm where a regressor is trained against the output actions to perform the task, conditioned by a single demonstration sampled from a given task. This approach is close to previous regression-based technique presented in Section~\ref{param_adaptation}, but formalised on a set of tasks.
For example, tasks can be  training the robot-arm of stacking a variable number of physical blocks among a variable number of piles. The evaluation methods rely on tests on seen and unseen demonstration during training. Their results showed that the robot performed equally well with seen and unseen demonstrations.

Adaptation process through meta-learning in motor learning has also been investigated with the model-agnostic meta-learning (MAML) method~\cite{finn2017model},
allowing faster weight adaptation to new examples representing a task. \cite{finn2017one, yu2018one} extended the MAML approach for one-shot imitation learning by a robotic arm. \cite{finn2017one} first demonstrate that vision-based policies can be fine-tuned from one demonstration. They conducted experiments using two types of tasks (pushing object and placing object in a recipient) on both a simulated and a real robot using video-based input data. Their results outperformed previous results (see for instance \cite{duan2017one}) in terms of the number of demonstrations needed for adaptation. Then, \cite{yu2018one} addressed the problem of one-shot learning of motor control policies with domain shift. Their experiments on simulated and real robot actions showed good results on tasks such as push, place, pick-and-place objects.

The MAML method has also found applications in human motion forecasting \cite{gui2018few}, for which large annotated motion capture are typically needed. They propose an approach based combining MAML and model regression networks~\cite{wang2017learning, wang2016learning}, allowing for learning a good generic initial model and for adaptation efficiently to unseen tasks. They showed that the model outperforms baselines with 5 examples of motion capture data of walking.

\section{Adaptation through reinforcement learning}
\label{RL}

Reinforcement Learning (RL) enables robotic agents to acquire new motor skills from experience, using trial-and-error interactions with its environment~\cite{kober2013reinforcement}. Contrary to the imitation learning approaches discussed in section~\ref{param_adaptation}, where expert demonstrations are used to train a model encoding a given behavior, RL relies on objective functions that provide feedback on the robot's performance.

Most approaches to imitation learning rely on a supervised paradigm where the model is fully specified from demonstrations without subsequent self-improvement~\cite{billard2016learning}.
To ensure a good task generalization, imitation learning requires a significant number of high quality demonstrations that provide variability while ensuring high performance. While RL can raise impressive performance, the learning process is often very slow and can lead to unnatural behavior. A growing body of research investigates the combination of these two paradigms to improve the models' adaptation to new tasks, making the learning process more efficient and improving the generalization of the tasks from few examples.

Demonstrations can be integrated in the RL process in various ways. One approach consists of initializing RL training with a model learned by imitation~\cite{kober2013reinforcement}, typically by a human teacher. Demonstrations of such tasks are used to generate initial policies for the RL process, enabling robots to rapidly learn to perform tasks such as reaching tasks, ball-in-a-cup, playing pool, manipulating a box, etc. A second strategy consists of deriving cost functions from demonstrations, for instance using inverse reinforcement learning~\cite{kolter2008hierarchical,finn2016guided}.
\cite{finn2016guided} showed that using 25-30 human demonstrations (by direct manipulation) to learn the cost function was sufficient for the robot to learn how to perform dish placement and pouring tasks.

Building upon the success of Generative Adversarial Networks in other fields of machine learning, Generative Adversarial Imitation Learning (GAIL) has been proposed as an efficient method for learning movement from expert demonstrations~\cite{ho2016generative}. In GAIL, a discriminator is trained to discriminate between expert trajectories (considered optimal) and policy trajectories generated by a generator that is trained to fool the discriminator. This approach was then extended to reinforcement learning through self-imitation where optimal trajectories are defined by previous successful attempts~\cite{guo2018generative}. Several extensions of the adversarial learning framework were proposed to improve its stability or to handle unstructured demonstrations~\cite{hausman2017multi,wang2017robust}.
These recent approaches have been evaluated on a standard set of tasks using simulated environments, in particular OpenAI Gym MuJoCo~\cite{todorov2012mujoco}, including continuous control tasks such as inverted pendulums, 4-legged walk, humanoid walk.

Recently, \cite{zhu2018reinforcement} proposed simultaneous imitation and reinforcement learning through a reward function that combines GAIL and RL. \cite{zhu2018reinforcement} evaluated their approach on several manipulation tasks (such as block lifting and stacking, clearing a table, pouring content) with a robot arm. Demonstrations were performed using a 3D controller, the training was done in a simulated environment, and the tasks were performed by a real robot arm. In comparison with GAIL or RL alone, the evaluation shows that the combination learns faster and reaches better performance.

\section{Discussion}
\label{sec:discussion}

This paper reviews three types of adaptation in machine learning applied to movement modeling. In this section, we discuss how adaptive movement models can be used to support motor learning, including both motor adaptation and motor skill acquisition.

First, motor adaptation mechanisms involve variations of an already-trained skill. Computationally, motor adaptation can be seen as an optimization process that learns and cancels external effects in order to return to baseline \cite{shadmehr1994adaptive}. Accounting for these underlying variations require rapid mechanisms and robust statistical modeling. Probabilistic model parameter adaptation (Section~\ref{param_adaptation}) appears to be a good candidate to understand movement variability induced by motor adaptation processes. However, while motor adaptation has been widely studied, very few is known on the statistical structure of motor adaptation, in particular trial-to-trial motor variability \cite{stergiou2011human}. Transfer learning could also be used: pre-trained models (RNNs or CNNs) that capture some structure of movement parameters (i.e. low-dimensional subspaces of the parameters space), can be adapted online for fine-grained variations. Here, open questions concern how such variations can account for structural learning in motor control \cite{braun2009motor}.

Second, more dramatic changes in movement patterns, as induced by learning new motor skills, might require computational adaptation that involves re-training procedures. Transfer and meta-learning (Section~\ref{transfer_meta_learning}) describe the adaptation of high-capacity movement models to new tasks, and could be used in this context. One difficulty is to assess to what extent transferring a given model to new motor control policies would induce the model to forget past skills. For instance, it was found that movement models relying on deep neural networks might lead to catastrophic forgetting \cite{kirkpatrick2017overcoming}. Also, meta-learning algorithms such as MAML \cite{finn2017model} are currently not suitable to adapt to several new motor tasks.
Self-imitation and reinforcement mechanisms (Section~\ref{RL}) could help to generalize to a wider set of tasks. A current challenge is to learn suitable action selection policies. Although exploration-exploitation is known to be central in motor learning \cite{herzfeld2014motor}, it is yet unclear what process drives action selection in the brain \cite{carland2019urge,sugiyama2020reinforcement}. These approaches still need to be experimentally assessed in a motor learning context.

Finally, a last challenge that we want to raise in this paper regards the continuous evolution of motor variation patterns. Motor execution may continuously vary over time, due to skill acquisition and morphological changes. Accounting for such open-ended task may require new form of adaptation such as continuous online learning, as proposed by \cite{nagabandi2018deep}. We think that this is a promising research direction, raising the central question of computation and memory in motor learning \cite{herzfeld2014memory}.

In closing, to be integrated in motor learning support systems, the aforementioned machine learning approaches should be combined with adaptation mechanisms that aim to generalize models to new movements and new tasks efficiently. We do not advocate solely for adaptive machine learning explaining motor learning processes. We propose adaptation procedures that can account for variation patterns observed in behavioural data, leading to performance improvements in motor learning support systems.

\section*{Conflict of Interest Statement}

The authors declare that the research was conducted in the absence of any commercial or financial relationships that could be construed as a potential conflict of interest.

\section*{Author Contributions}

BC made the first draft and all authors contributed to the manuscript, adding content and revising it critically for important intellectual content.

\section*{Funding}

This research was supported by the ELEMENT project (ANR-18-CE33-0002) and the ARCOL project (ANR-19-CE33-0001) from the French National Research Agency.

\bibliographystyle{unsrt}

\end{document}